\documentclass[twoside,11pt]{article}

\usepackage{jmlr2e}
\usepackage{amsmath}
\usepackage{times}
\usepackage{booktabs}

\usepackage{multirow, makecell}
\usepackage{rotating}
\usepackage{subfig}
\usepackage{caption}
\usepackage{graphicx}
\usepackage[shortlabels]{enumitem}
\ShortHeadings{Clinical Event Prediction and Understanding using Deep Networks}{Suresh, Hunt, Johnson, Celi, Szolovits and Ghassemi}%{Lastname, PhD and Lastname, MD}
\firstpageno{1}

\begin{document}

\title{Clinical Intervention Prediction and Understanding using Deep Networks}

\author{\name Harini Suresh \email hsuresh@mit.edu \\
       \AND
       \name Nathan Hunt \email  nhunt@mit.edu\\
       \AND
       \name Alistair Johnson \email aewj@mit.edu\\
       \AND
       \name Leo Anthony Celi \email lceli@mit.edu \\
       \AND
       \name Peter Szolovits \email psz@mit.edu 		\\
       \AND
       \name Marzyeh Ghassemi \email mghassem@mit.edu \\
       \\
       \addr Computer Science and Artificial Intelligence Lab, MIT\\
       Cambridge, MA\\}

\maketitle

%TODO: UPDATE THE ABSTRACT!
\begin{abstract}
Real-time prediction of clinical interventions remains a challenge within intensive care units (ICUs). This task is complicated by data sources that are noisy, sparse, heterogeneous and outcomes that are imbalanced. In this paper, we integrate data from all available ICU sources (vitals, labs, notes, demographics) and focus on learning rich representations of this data to predict onset and weaning of multiple invasive interventions. In particular, we compare both long short-term memory networks (LSTM) and convolutional neural networks (CNN) for prediction of five intervention tasks: invasive ventilation, non-invasive ventilation, vasopressors, colloid boluses, and crystalloid boluses.
Our predictions are done in a forward-facing manner to enable ``real-time'' performance, and predictions are made with a six hour gap time to support clinically actionable planning. We achieve state-of-the-art results on our predictive tasks using deep architectures. We explore the use of feature occlusion to interpret LSTM models, and compare this to the interpretability gained from examining inputs that maximally activate CNN outputs. We show that our models are able to significantly outperform baselines in intervention prediction, and provide insight into model learning, which is crucial for the adoption of such models in practice.
\end{abstract}

\section{Introduction}
As Intensive Care Units (ICUs) play an increasing role in acute healthcare delivery \citep{vincent2013critical}, clinicians must anticipate patient care needs in a fast-paced, data-overloaded setting. The secondary analysis of healthcare data is a critical step toward improving modern healthcare, as it affords the study of care in the real care settings and patient populations. 
%In order to provide clinical staff with actionable recommendations for patient care, we plan to gain insight from healthcare data in Electronic Health Records (EHR). EHR data is collected for the primary purpose of facilitating patient care. EHR systems that meet federal requirements are present in most acute care hospitals (97\% in 2014 \citep{charles2013adoption}) and office-based physicians (78\% in 2015 \citep{jamoom2015prevalence}). 
The widespread availability of electronic healthcare data \citep{charles2013adoption,jamoom2015prevalence} allows new investigations into evidence-based decision support, where we can learn when patients need a given intervention. Continuous, forward-facing event prediction is particularly applicable in the ICU setting where we want to account for evolving clinical needs and information throughout the patient's stay.

In this work, we focus on predicting the onset and weaning of interventions. The efficacy of interventions can vary drastically from patient to patient, and unnecessarily administering an intervention can be harmful and expensive. Any treatments come with inherent risks, and we target interventions that span a wide severity of needs in critical care--- specifically, invasive ventilation, non-invasive ventilation, vasopressors, colloid boluses, and crystalloid boluses. Mechanical ventilation is commonly used for breathing assistance, but has many potential complications \citep{yang1991prospective} and small changes in ventilation settings can have large impact in patient outcomes \citep{tobin2006principles}. Vasopressors are a common ICU medication, but there is no robust evidence of improved outcomes from their use \citep{mullner2004vasopressors}, and some evidence they may be harmful \citep{d2015blood}. Fluid boluses are used to improve cardiovascular function and organ perfusion. There are two bolus types: crystalloid and colloid. Both are often considered as less aggressive alternatives to vasopressors, but there are no multi-center trials studying whether fluid bolus therapy should be given to critically ill patients, only studies trying to distinguish which type of fluid should be given \citep{malbrain2014fluid}.

Capturing complex relationships across many disparate data types is key for predictive performance in our tasks. To this end, we take advantage of the success of deep learning models to capture rich representations of data with little hand-engineering by domain experts. We use long short-term memory networks (LSTM) \citep{hochreiter1997long}, which have been shown to effectively model complicated dependencies in timeseries data \citep{bengio1994learning}. Previously, LSTMs have achieved state-of-the-art results in many different applications, such as machine translation \citep{hermann2015teaching}, dialogue systems \citep{chorowski2015attention} and image captioning \citep{xu2015show}. They are well-suited to our modeling tasks because clinical conditions may be spread apart over several hours. We compare the LSTM models to a convolutional neural network (CNN) architecture that has previously been explored for longitudinal laboratory data \citep{razavian2016multi}. All models predict outcomes in a continuous manner given any patient record over vitals, labs, demographic, and notes. In doing so, we:
\begin{enumerate}[nosep]
\item Achieve state-of-the-art prediction results in our forward-facing, hourly prediction of clinical interventions (onset, weaning, and continuity) that could be used at the time of care.
\item Demonstrate that different data modalities and features are most important for different types of predictive tasks in our LSTM using feature occlusion. This is an important step in making models more interpretable by physicians.
\item Highlight patient trajectories that lead to the most and least confident predictions in our CNN across outcomes and features, also aiding in interpretability.
\end{enumerate}

\section{Background and Related Work}
Clinical decision-making often happens in settings of limited knowledge and high uncertainty; for example, only 10 of the 72 ICU interventions evaluated in randomized controlled trials (RCTs) are not associated with improved outcomes \citep{ospina2008multicenter}. Our goal is to gain insight from healthcare data previously collected for the primary purpose of facilitating patient care.

Recent studies have applied recurrent neural networks (RNNs) to modeling sequential EHR data to tag ICU signals with billing code labels \citep{che2016recurrent,lipton2015learning, DBLP:journals/corr/ChoiBS15}, to identify the impact of different drugs for diabetes \citep{krishnan2015deepkalman}. \cite{razavian2016multi} compared CNNs to LSTMs for longitudinal outcome prediction on billing codes using lab tests. With regard to interpretability, \cite{choi2016retain} used temporal attention to identify important features in early diagnostic prediction of chronic diseases from time-ordered billing codes. Others have focused on using representations of clinical notes \citep{ghassemi2014unfolding} or patient physiological signals to predict mortality \citep{ghassemi2015multivariate}.  

Previous work on interventions in ICU populations have often either focused on a single outcome or used data from specialized cohorts. Such models with vasopressors as a predictive target have achieved AUCs of 0.79 in patients receiving fluid resuscitation \citep{fialho2013disease}, 0.85 in septic shock patients \citep{salgado2016ensemble}, and 0.88 for onset after a 4 hour gap and 0.71 for weaning, only trained on patients who did receive a vasopressor \citep{wu2016ssam}. However, we train our models on general ICU populations in order to make them more applicable. In the most recent prior work on interventions, also on a general ICU population, the best AUC performances were 0.67 (ventilation), 0.78 (vasopressor) for vasopressor onset prediction after a 4 hour gap \citep{ghassemi2017predicting}. These were lowered to 0.66 and 0.74 with a longer gap time of 8 hours.

\section{Data and Preprocessing }
\begin{figure*}[t!]
\centering
\includegraphics[width=\linewidth]{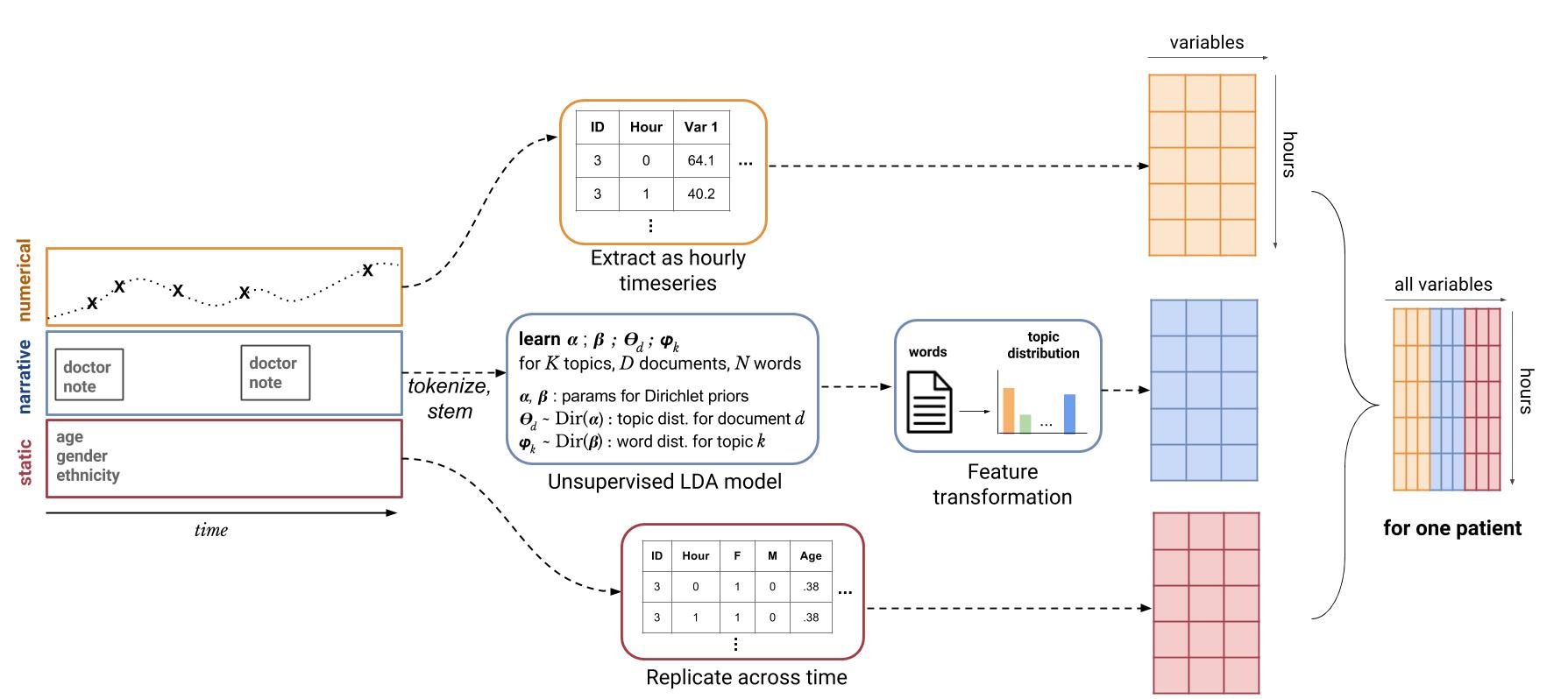}
\caption{Data preprocessing and feature extraction with numerical measurements and lab values, clinical notes and static demographics.}
\label{fig:experimentalFlow}
\end{figure*}

See Figure \ref{fig:experimentalFlow} for an overall description of data flow.

\subsection{Data Source}
We use data from the Multiparameter Intelligent Monitoring in Intensive Care (MIMIC-III v1.4) database \citep{johnson2016mimiciii}. MIMIC is publicly available, and contains over 58,000 hospital admissions from approximately 38,600 adults. We consider patients 15 and older who had ICU stays from 12 to 240 hours and consider each patient's first ICU stay only. This yields 34,148 unique ICU stays.

\subsection{Data Extraction and Preprocessing}
For each patient, we extract:
\begin{enumerate}[nosep]
\item 5 static variables such as gender and age
\item 29 time-varying vitals and labs such as oxygen saturation and blood urea nitrogen
\item All available, de-identified clinical notes for each patient as timeseries across their entire stay 
\end{enumerate}
(See Table \ref{tab:variables} for a complete listing of variables)

Static variables were replicated across all timesteps for each patient. Vital and lab measurements are given timestamps that are rounded to the nearest hour. If an hour has multiple measurements for a signal, those measurements are averaged. 

\subsection{Representation of Notes and Vitals}
\label{sec:rep_physwords}
Clinical narrative notes were processed to create a $50$-dimensional vector of topic proportions for each note using Latent Dirichlet Allocation \citep{LDA,GrifStey}. These vectors are replicated forward and aggregated through time \citep{ghassemi2014unfolding}. For example, if a patient had a note $A$ recorded at hour 3 and a note $B$ at hour 7, hours 3--6 would contain the topic distribution from $A$, while hours 7 onward would contain the aggregated topic distribution from $A$ and $B$ combined.

We compare raw physiological data to \emph{physiological words}, where we categorize the vitals data by first converting each value into a z-score based on the population mean and standard deviation for that variable, and then rounding this score to the nearest integer and capping it to be between -4 and 4. Each z-score value then becomes its own column, which explicitly allows for a representation of missingness (e.g., all columns for a particular variable zeroed) that does not require imputation (Figure \ref{fig:physWords} in Appendix B) \citep{wu2016ssam}.

The physiological variables, topic distribution, and static variables for each patient are concatenated into a single feature vector per patient per hour \citep{esteban2016predicting}. The intervention state of each patient (a binary value indicating whether or not they are on the intervention of interest at each timestep) and the time of day for each timestep (an integer from 0 to 23 representing the hour) are also added to this feature vector. Using the time of day as a feature makes it easier for the model to capture circadian rhythms that may be present in, e.g., the vitals data.

\subsection{Prediction Task}
\begin{figure*}[t!]
\centering
\includegraphics[width=.5\linewidth]{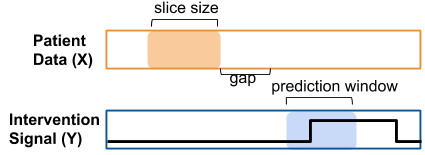}
\caption{Given data from a fixed-length (6 hour) sliding window, models predict the status of intervention in a prediction window (4 hours) after a gap time (6 hours). Windows slide along the entire patient record, creating multiple examples from each record.}
\label{fig:task}
\end{figure*}

We split each patient’s record into 6 hour chunks using a sliding window and make a prediction for a window of 4 hours after a gap time of 6 hours (Figure \ref{fig:task}). 
When predicting ventilation, non-invasive ventilation, or vasopressors, the model classifies the prediction window as one of four possible outcomes: 1) Onset, 2) Wean, 3) Staying on intervention, 4) Staying off intervention. A prediction window is an onset if there is a transition from a label of 0 to 1 for the patient during that window; weaning is the opposite: a transition from 1 to 0. A window is classified as "stay on" if the label for the entire window is 1 or "stay off" if 0. When predicting colloid or crystalloid boluses, we classify the prediction window into one of two classes: 1) Onset, or 2) No Onset, since these interventions are not administered for on-going durations of time. 
After splitting the patient records into fixed-length chunks, we end up with 1,154,101 examples. Table \ref{tab:proportions} lists the proportions of each class for each intervention.
\begin{table}
\centering
\begin{tabular}{ | c || c | c | c | c |}
  \hline
  & Onset & Weaning & Stay Off & Stay On \\  \hline			
  Ventilation & 0.005 & 0.017 & 0.798 & 0.18\\
  Vasopressor & 0.008 & 0.016 & 0.862 & 0.114\\
  NI-Ventilation & 0.024 & 0.035 & 0.695 & 0.246 \\
  Colloid Bolus & 0.003 & - & - & -\\
  Crystalloid Bol & 0.022 & - & - & -\\
  \hline  
\end{tabular}
\caption{The proportion of each intervention class. Note that colloid and crystalloid boluses are not administered for specific durations, and thus have only a single class (onset). NI = non-invasive.}
\label{tab:proportions}
\end{table}

\section{Methods}
\subsection{Long Short-Term Memory Network (LSTM)}
We use long short-term memory networks (LSTM) as our first model. Having seen the input sequence $x_1 \ldots x_{t}$ of a given example, we predict $\hat{y_t}$, a probability distribution over the outcomes, with target outcome $y_t$:
\begin{align}
h_1 \ldots h_t =  \textsl{LSTM}(x_1 \ldots x_{t}) \\
\hat{y_t} =  \textsl{softmax}(W_yh_t + b_y)
\end{align}
\noindent  where $x_i \in \mathbb{R}^{V}, W_y \in \mathbb{R}^{N_C \times L_2}, h_t \in \mathbb{R}^{L_2}$, $b_y \in \mathbb{R}^{N_C}$ where $V$ is the dimensionality of the input (number of variables), $N_C$ is the number of classes we predict, and $L_2$ is the second hidden layer size. For a model schematic, see Figure \ref{fig:lstm}, and for more details on model implementation, see the Appendix.

\subsection{Convolution Neural Network (CNN)}
We employ a similar CNN architecture to \cite{razavian2016multi}, except that we do not initially convolve the features into an intermediate representation. We represent features as channels and perform 1D temporal convolutions, rather than treating the input as a 2D image. Our architecture consists of temporal convolutions at three different temporal granularities with 64 filters each. The dimensions of the filters are $1 \times i$, where $i \in \{3,4,5\}$. 

\begin{figure}[htbp]
\centering
\subfloat[The LSTM consists of two hidden layers with 512 nodes each. We sequentially feed in each hour's data. At the end of the example window, we use the final hidden state to predict the output. \label{fig:lstm}]{\includegraphics[width=.44\textwidth]{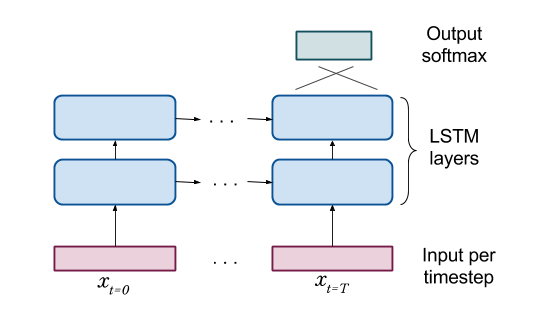}}
\hspace{0.05\textwidth}
\subfloat[The CNN architecture performs temporal convolutions at 3 different granularities (3, 4, and 5 hours), max-pools and combines the outputs, and runs this through 2 fully connected layers to arrive at the prediction.\label{fig:cnn}] {\includegraphics[width=.48\textwidth]{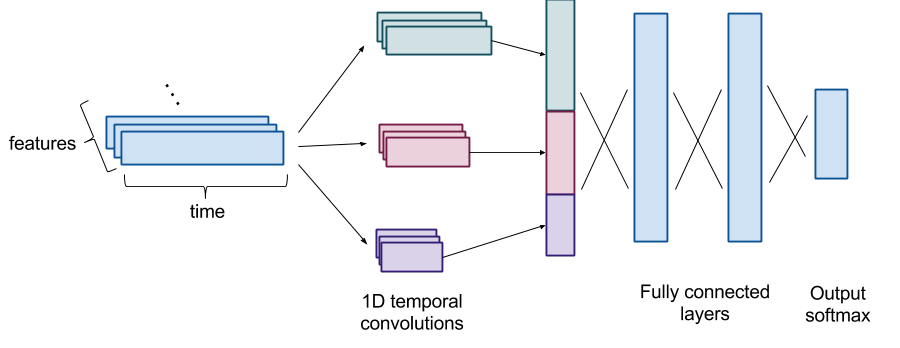}}\hfill
\caption{Schematics of LSTM and CNN model architectures.} \label{fig:models}
\end{figure}

We pad the inputs such that the outputs from the convolutional layers are the same size, and we use a stride of 1. Each convolution is followed by a max pooling layer with a pooling size of 3. The outputs from all three temporal granularities are concatenated and flattened, and followed by 2 fully connected layers with dropout in between and a softmax over the output (Figure \ref{fig:cnn}).

\subsection{Experimental Settings}
We use a train/validation/test split of 70/10/20 and stratify the splits based on outcome.  For the LSTM, we use dropout with a keep probability of 0.8 during training (only on stacked layers), and L2 regularization with lambda = 0.0001. We use 2 hidden LSTM layers of 512 nodes each.  For the CNN, we use dropout between fully-connected layers with a keep probability of 0.5. 
We use a weighted loss function during optimization to account for class imbalances.
All parameters were determined using cross-validation with the validation set. We implemented all models in TensorFlow version 1.0.1 using the Adam optimizer on mini-batches of 128 examples. We determine when to stop training with early stopping based on the AUC on the validation set.

\subsection{Evaluation} 
We evaluate our results based on per-class AUCs as well as aggregated macro AUCs. If there are $K$ classes each with a per-class AUC of $AUC_k$ then the macro AUC is defined as the average of the per-class AUCS, $AUC_{macro}=\frac{1}{K}\sum_k{AUC_k}$. We use the macro AUC as an aggregate score because it weights the AUCs of all classes equally, regardless of class size \citep{macro_auc}. This is important because of the large class imbalance present in the data.

We use L2 regularized logistic regression (LR) as a baseline for comparison with the neural networks \citep{scikit-learn}. The same input is used as for the numerical LSTM and CNN (imputed 6 hour chunks of data).

\subsection{Interpretibility}
\subsubsection{LSTM Feature-Level Occlusions}
Because of the additional time dependencies of recurrent neural networks, getting feature-level interpretability from LSTMs is notoriously difficult. To achieve this, we borrow an idea from image recognition to help understand how the LSTM uses different features of the patients. \cite{DBLP:journals/corr/ZeilerF13} use occlusion to understand how models process images: they remove a region of the image (by setting all values in that region to 0) and compare the model's prediction of this occluded image with the original prediction. A large shift in the prediction implies that the occluded region contains important information for the correct prediction. With our LSTM model, we remove features one by one from the patients (by replacing the given feature with noise drawn from a uniform distribution in [0,1)). We then compare the predictive ability of the model with and without each feature; when this difference is large, then the model was relying heavily on that feature to make the prediction. 

\subsubsection{CNN Filter/Activation Visualization}
We get interpretability from the CNN models in two ways. First, in order to understand how the CNN is using the patient data to predict certain tasks, we find and compare the top 10 real examples that our model predicts are most and least likely to have a specific outcome. As our gap time is 6 hours, this means that the model predicts high probability of onset of the given task 6 hours after the end of the identified trajectories.

Second, we generate ``hallucinations'' from the model which maximize the predicted probability for a given task  \citep{erhan2009visualizing}. This is done by creating an objective function that maximizes the activation of a specific output node, and backpropagating gradients back to the input image, adjusting the image so that it maximally activates the output node.  

\section{Results} 
We found deep architectures achieved state-of-the-art prediction results for our intervention tasks. The AUCs for each of our five intervention types and 4 prediction tasks are shown for all models in Table \ref{tab:AUCs}. All models use 6 hour chucks of ``raw'' data which have either been transformed to a 0-1 range (normalized and mean imputed), or discretized into physiological words (Section \ref{sec:rep_physwords}).

\subsection{Physiological Words Improve Predictive Task Performance With High Class Imbalance}
We observed a significantly increased AUC for some interventions when we used physiological words --- specifically for ventilation onset (from 0.61 to 0.75) and colloid bolus onset (from 0.52 to 0.72), which have the lowest proportion of onset examples (Table \ref{tab:proportions}). This may be because physiological words have a smoothing effect. Since we round the z-score for each value to the nearest integer, if a patient has a heart rate of 87 at one hour and then 89 at the next, those will probably be represented as the same word. This may make the model invariant to small fluctuations in the patient's data and more resilient to overfitting small classes. 
In addition, the physiological word representation has an explicit encoding for missing data. This is in contrast to the raw data that has been forward-filled and mean-imputed, introducing noise and making it difficult for the model to know how confident to be in the measurements it is given \citep{che2016recurrent}. 

% TODO: give a statistic about how much data is missing?

% TODO: compare using the same task to AMIA CRI paper Marzyeh's; how do our results compare?

\begin{table}[htb]
\newcommand{\rot}[1]{\rotatebox{90}{#1}}
\newcommand{\multirot}[1]{\multirow{4}{*}[1.5ex]{\rotcell{#1}}}
 \centering 
 \begin{tabular}{c|c||c|c|c|c|c||}
 	& & \multicolumn{5}{c||}{\textbf{Intervention Type}}\\
    %\cmidrule{3-7}
	\textbf{Task} & \textbf{Model} & VENT & NI-VENT & VASO & COL BOL & CRYS BOL\\
    \hline
    \multirot{Onset AUC} & Baseline & 0.60 & 0.66 & 0.43 & 0.65 & 0.67 \\
    & LSTM Raw & 0.61 & 0.75 & \textbf{0.77} & 0.52 & 0.70 \\
    & LSTM Words & \textbf{0.75} & \textbf{0.76} & 0.76 & \textbf{0.72} & \textbf{0.71} \\
    & CNN & 0.62 & 0.73 & \textbf{0.77} & 0.70 & 0.69 \\
    \hline
    \multirot{Wean AUC} & Baseline & 0.83 & 0.71 & 0.74 & - & - \\
    & LSTM Raw & 0.90 & 0.80 & \textbf{0.91} & - & - \\
    & LSTM Words & 0.90 & \textbf{0.81} & \textbf{0.91} & - & - \\
    & CNN & \textbf{0.91} & 0.80 & \textbf{0.91} & - & - \\
    \hline
    \multirot{Stay On AUC} & Baseline & 0.50 & 0.79 & 0.55 & - & - \\
    & LSTM Raw & 0.96 & \textbf{0.86} & \textbf{0.96} & - & - \\
    & LSTM Words & \textbf{0.97} & \textbf{0.86} & 0.95 & - & - \\
    & CNN & 0.96 & \textbf{0.86} & \textbf{0.96} & - & - \\
    \hline
    \multirot{Stay Off AUC} & Baseline & 0.94 & 0.71 & 0.93 & - & - \\
    & LSTM Raw & 0.95 & \textbf{0.86} & \textbf{0.96} & - & - \\
    & LSTM Words & \textbf{0.97} & \textbf{0.86} & 0.95 & - & - \\
    & CNN & 0.95 & \textbf{0.86} & \textbf{0.96} & - & - \\
    \hline
    \\
    \hline
    \multirot{Macro AUC} & Baseline & 0.72 & 0.72 & 0.66 & - & - \\
    & LSTM Raw & 0.86 & \textbf{0.82} & \textbf{0.90} & - & - \\
    & LSTM Words & \textbf{0.90} & \textbf{0.82} & 0.89 & - & - \\
    & CNN & 0.86 & 0.81 & \textbf{0.90} & - & - \\
 \end{tabular}
  \caption{Comparison of model performance on five targeted interventions. Models that perform best for a given (intervention, task) pair are bolded.} 
  \label{tab:AUCs}
\end{table}

\subsection{Feature-Level Occlusions Identify Important Per-Class Features}
\begin{figure}[ht]
\centering
\subfloat[\label{fig:vent-onset}] {\includegraphics[width=.23\textwidth]{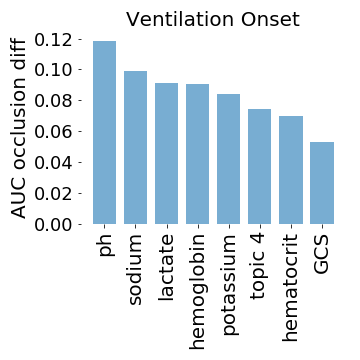}}
~
\subfloat[\label{fig:vent-wean}] {\includegraphics[width=.23\textwidth]{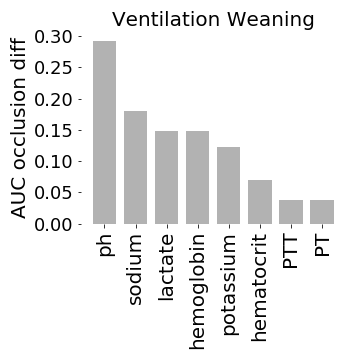}}
~
\subfloat[\label{fig:vaso-onset}]{\includegraphics[width=.23\textwidth]{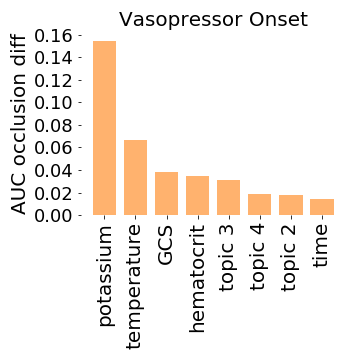}}
~
\subfloat[\label{fig:vaso-wean}] {\includegraphics[width=.23\textwidth]{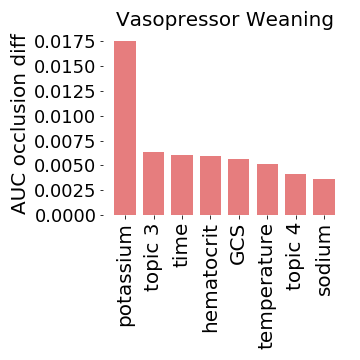}}

\subfloat[\label{fig:niv-onset}]{\includegraphics[width=.23\textwidth]{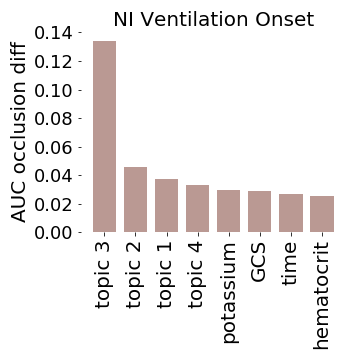}}
~
\subfloat[\label{fig:niv-wean}] {\includegraphics[width=.23\textwidth]{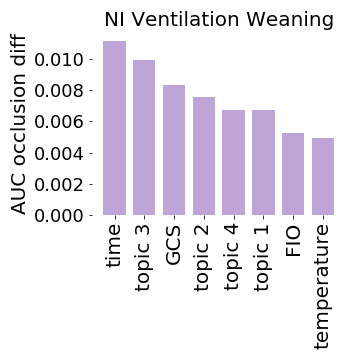}}
~
\subfloat[\label{fig:crys-onset}]{\includegraphics[width=.23\textwidth]{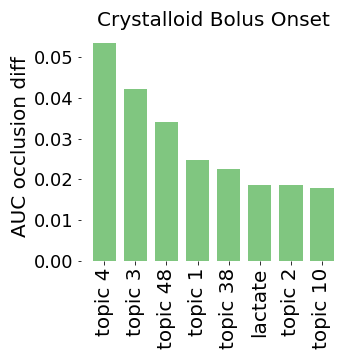}}
~
\subfloat[\label{fig:col-onset}]{\includegraphics[width=.23\textwidth]{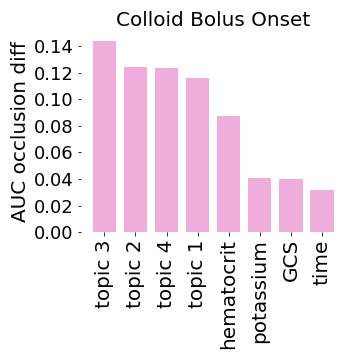}}

\caption{We are able to make interpretable predictions using the LSTM and occluding specific features. The top eight features that cause a decrease in prediction AUC for each intervention task. In general, physiological data were more important for the more invasive interventions --- mechanical ventilation (\ref{fig:vent-onset}, \ref{fig:vent-wean}) and vasopressors (\ref{fig:vaso-onset}, \ref{fig:vaso-wean}) --- while clinical note topics were more important for less invasive tasks --- non-invasive ventilation (\ref{fig:niv-onset}, \ref{fig:niv-wean}) and fluid boluses (\ref{fig:crys-onset}, \ref{fig:col-onset}). Note that all weaning tasks except for ventilation have significantly less AUC variance.} \label{fig:occlusions}
\end{figure}

We are able to interpret the LSTM's predictions using feature occlusion (Section 4.5.1). We note that vitals, labs, topics and static data are important for different interventions (Figure \ref{fig:occlusions}). Table \ref{tab:top_words} has a complete listing of the most probable words for each topic mentioned.

For mechanical ventilation, the top five important features are consistent for weaning and onset (pH, sodium, lactate, hemoglobin, and potassium). This is sensible, because all are important lab values used to assess a patient's physiological stability, and ventilation is an aggressive intervention. However, ventilation onset additionally places importance on a patient's Glasgow Coma Score (GCS) and Topic 4 (assessing patient consciousness), likely because patient sedation is a critical part of mechanical ventilation. We also note that the scale of AUC difference between ventilation onset and weaning is the largest observed (up to 0.30 for weaning and 0.12 for onset). 

In vasopressor onset prediction, physiological variables such as potassium and hematocrit are consistently important, which agrees with clinical assessment of cardiovascular state \citep{bassi2013therapeutic}. Similarly, Topic 3 (noting many physiological values) is also important for both onset and weaning. Note that the overall difference in AUC for onset ranges up to 0.16, but there is no significant decrease in AUC for weaning ($<$ 0.02).  This is consistent with previous work that demonstrated weaning to be a more difficult task in general for vasopressors \citep{wu2016ssam}. We also note that weaning prediction places importance on time of day. As noted by \cite{wu2016ssam}, this could be a side-effect of patients being left on interventions longer than necessary.

For non-invasive ventilation onset and weaning the learned topics are more important than physiological variables. This may mean that the need for less severe interventions can only be detected from clinical insights derived in notes. Similarly to vasopressors, we note that onset AUCs vary more than weaning AUCs (0.14 vs 0.01), and that time of day is important for weaning. 

For crystalloid and colloid bolus onsets, topics are all but one of the five most important features for detection. Colloid boluses in general have more AUC variance for the topic features (0.14 vs. 0.05), which is likely due to the larger class imbalance compared to crystalloids. 

\subsection{Convolutional Filters Target Short-term Trajectories}
\begin{figure*}[t!]
\centering
\includegraphics[width=\textwidth]{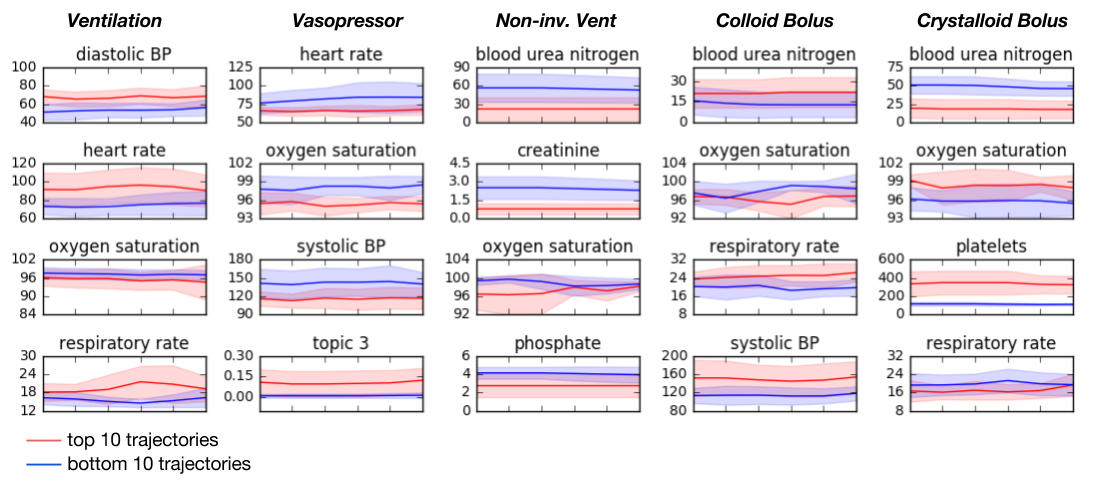}
\caption{Trajectories of the 10 maximally and minimally activating examples for onset of each of the interventions.}
\label{fig:trajectories}
\end{figure*}

We are able to understand the CNN by examining maximally activating patient trajectories (Section 4.5.2). Figure \ref{fig:trajectories} shows the mean with standard deviation error bars for four of the most differentiated features of the 10 real patient trajectories that are the highest and lowest activating for each task. The trends suggest that patients who will require ventilation in the future have higher diastolic blood pressure, respiratory rate, and heart rate, and lower oxygen saturation --- possibly corresponding to patients who are experiencing hyperventilation. For vasopressor onsets, we see a decreased systolic blood pressure, heart rate and oxygen saturation rate. These could either indicate altered peripheral perfusion or stress hyperglycemia. Topic 3, which was important for vasopressor onset using occlusion \ref{fig:occlusions}, is also increased. 

In the less invasive tasks, we saw decreased creatinine, phosphate, oxygen saturation and blood urea nitrogen for non-invasive ventilation, potentially indicating neuromuscular respiratory failure. For colloid and crystalloid boluses we note general indicators of physiological decline, as boluses are given for a wide range of conditions.

\begin{figure*}[t!]
\centering
\includegraphics[width=\textwidth]{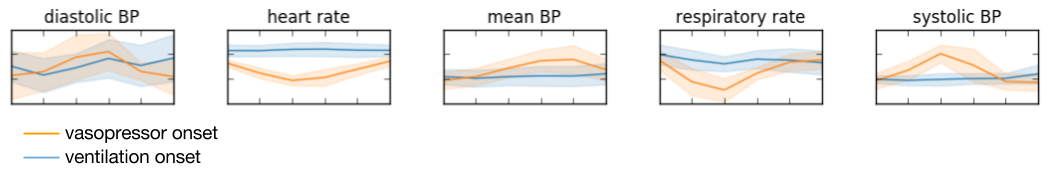}
\caption{Trajectories generated by adjusting inputs to maximally activate a specific output node of the CNN.}
\label{fig:hallucinations}
\end{figure*}

``Hallucinations'' for vasopressor and ventilation onset are shown in Figure \ref{fig:hallucinations}. While our model was not trained with any physiological knowledge or priors, we note that it identifies blood pressure drops as being maximally activating for vasopressor onset, and respiratory rate decreasing for ventilation onset. This suggests that it is still able to independently learn  physiological factors that are important for intervention prediction. We note that these hallucinations give us more insight into underlying properties of the network and what it is looking for. However, since these trajectories are made to maximize the output of the model, they do not necessarily correspond to physiologically plausible trajectories.

\section{Conclusion}
In this work, we targeted forward-facing prediction of ICU interventions covering multiple physiological organ systems. To our knowledge, our model is the first to use deep neural networks to predict both onset and weaning of interventions using all available modalities of ICU data. In our tasks, deep learning methods beat state-of-the-art AUCs reported in prior work for intervention prediction tasks --- this is sensible given that prior works have focused on single targets with smaller datasets \citep{wu2016ssam} or unsupervised representations prior to supervised training \citep{ghassemi2017predicting}. We also note that the LSTM over physiological words significantly improved performance in the two intervention tasks with the lowest incidence rate --- possibly because this representation encodes important information about what is ``normal'' for each physiological value, or is more robust to missingness in the physiological data.

Importantly, we were able to gain interpretability in both models. In the LSTMs, we examined feature importance using occlusion, and found that physiological data were important in more invasive tasks, while clinical note topics were more important for less invasive interventions. This could indicate that there is more clinical discretion at play for less invasive tasks. We also found that all weaning tasks save ventilation had less AUC variance, which could indicate that these decisions are also made with a large amount of clinical judgment. 

The temporal convolutions in our CNN filters over the multi-channel input learnt interesting and clinically-relevant trends in real patient trajectories, and these were further mimicked in the hallucinations generated by the network. As in prior work, we found that RNNs often have similar or improved performance as compared to CNNs \cite{razavian2016multi}. However, it is possible that more complex models would perform better as they uncover more long and short-term dependencies. 

Our results are an interesting start to extracting interpretability from neural networks on patient data, and future work to expand this will enable these models to be adopted in real clinical settings.

\acks{This research was funded in part by the Intel Science and Technology Center for Big Data and the National Library of Medicine Biomedical Informatics Research Training grant (NIH/NLM 2T15 LM007092-22). }

\bibliography{sample}

\clearpage
\appendix

\section*{Appendix}

\section{Dataset Statistics}

\begin{table}[h!]
\centering
\caption{Variables}
\begin{tabular}{c|ccc}
\hline
Static Variables & Gender & Age & Ethnicity \\
 & ICU & Admission Type\\
\hline
Vitals and Labs & Anion gap & Bicarbonate & blood pH\\
 & Blood urea nitrogen & Chloride & Creatinine\\
 & Diastolic blood pressure & Fraction inspired oxygen & Glascow coma scale total\\
 & Glucose & Heart rate & Hematocrit\\
 & Hemoglobin & INR\textsuperscript{*} & Lactate\\
 & Magnesium & Mean blood pressure & Oxygen saturation\\
 & Partial thromboplastin time & Phosphate & Platelets\\
 & Potassium & Prothrombin time & Respiratory rate\\
 & Sodium & Systolic blood pressure & Temperature\\
 & Weight & White blood cell count\\
\hline
\multicolumn{4}{l}{\textsuperscript{*}\footnotesize{International normalized ratio of the prothrombin time}}
\end{tabular}\label{tab:variables}
\end{table}
% \footnote{A crystalloid bolus is defined as giving a patient a crystalloid volume $ge$ 250 mL in 1 hour. A colloid bolus is a bit less well defined, but we approximate it with $\ge$ 100 mL in 1 hour. Colloid boluses are no longer used, but we include them as they do appear in the data.}

\begin{table}[h!]
\centering
\caption{Dataset Statistics}
\begin{tabular}{lll|l} 
~     & Train & Test & Total   \\
\hline
Patients & 27,318 & 6,830 & 34,148\\
Notes & 564,652 & 140,089 & 703,877\\
Elective Admission & 4,536 & 1,158 & 5,694\\
Urgent Admission & 746 & 188 & 934\\
Emergency Admission & 22,036 & 5,484 & 27,520\\
Mean Age & 63.9 & 64.1 & 63.9\\
Black/African American & 1,921 & 512 & 2,433\\
Hispanic/Latino & 702 & 166 & 868\\
White & 19,424 & 4,786 & 24,210\\
CCU (coronary care unit) & 4,156 & 993 & 5,149\\
CSRU (cardiac surgery recovery) & 5,625 & 1,408 & 7,033\\
MICU (medical ICU) & 9,580 & 2,494 & 12,074\\
SICU (surgical  ICU) & 4,384 & 1,074 & 5,458\\
TSICU (trauma SICU) & 3,573 & 861 & 4,434\\
Female & 11,918 & 2,924 & 14,842\\
Male & 15,400 & 3,906 & 19,306\\
ICU Mortalities & 1,741 & 439 & 2,180\\
In-hospital Mortalities & 2,569 & 642 & 3,211\\
30 Day Mortalities & 2,605 & 656 & 3,216\\
90 Day Mortalities & 2,835 & 722 & 3,557\\
Vasopressor Usage & 8,347 & 2,069 & 10,416\\
Ventilator Usage & 11,096 & 2,732 & 13,828\\
\end{tabular}\label{tab:statistics} 
\end{table}

\section{Physiological Word Generation}
\begin{figure*}[h!]
\centering
\includegraphics[width=\linewidth]{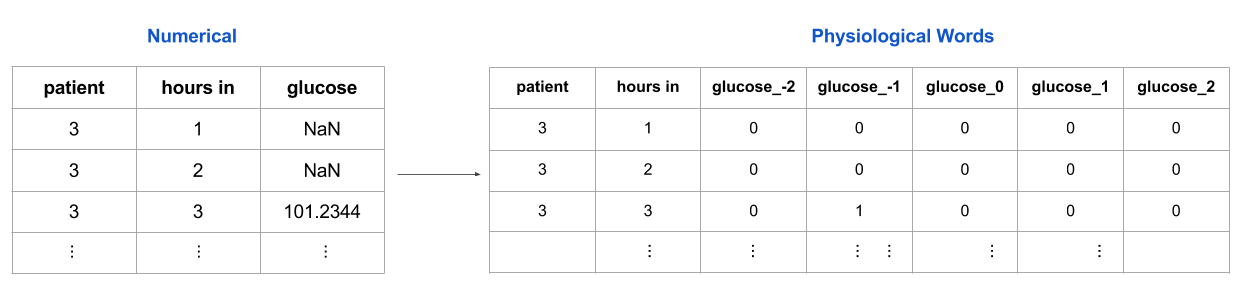}
\caption{Converting data from continuous timeseries format to discrete ``physiological words.'' The numeric values are first z-scored and rounded, and then each z-score is made into its own category. On the right, glucose\_-2 indicates the presence of a glucose value that was 2 standard deviations below the mean. A row containing all zeros for a given variable indicates that the value for that variable was missing at the timestep.}
\label{fig:physWords}
\end{figure*}

See Figure \ref{fig:physWords}.

\section{LSTM Model Details}
\textit{LSTM} performs the following update equations for a single layer, given its previous hidden state and the new input:
\begin{align}
f_t = \sigma(W_f[h_{t-1}, x_t] + b_f) \\
i_t = \sigma(W_i[h_{t-1}, x_t] + b_i)\\
\tilde{c_t} = \text{tanh}(W_c[h_{t-1}, x_t] + b_c)\\
c_t = f_t \odot c_{t-1} + i_t \odot \tilde{c_t}\\
o_t = \sigma(W_o[h_{t-1}, x_t] + b_o)\\
h_t = o_t \odot tanh(c_t)
\end{align}

\noindent where $W_f, W_i, W_c, W_o \in \mathbb{R}^{L_1 \times (L_1 + V)}$, $b_f, b_i, b_c, b_o \in \mathbb{R}^{L_1}$ are learned parameters, and $f_t, i_t, \tilde{c_t},$ $c_t, o_t, h_t \in \mathbb{R}^{L_1}$. In these equations, $\sigma$ stands for an element-wise application of the sigmoid (logistic) function, and $\odot$ is an element-wise product.
This is generalized to multiple layers by providing $h_t$ from the previous layer in place of the input.

We calculate classification loss using categorical cross-entropy, which sets the loss for predictions for $N$ examples over $M$ classes as:
$$\mathcal{L}(\hat{y}_1 \ldots \hat{y}_N) = -\frac{1}{N} \sum\limits^{N}_{i=1} \sum\limits^{M}_{j=1} y_{ij} \log\hat{y}_{ij}$$

\noindent where $\hat{y}_{ij}$ is the probability our model predicts for example $i$ being in class $j$, and ${y}_{ij}$ is the true value.

\section{Generated Topics}
\begin{table}[h!]
\centering
\caption{Most probable words in the topics most important for intervention predictions.}
\begin{tabular}{c|p{9cm}|p{3cm}} 
\textbf{Topic} & \centering{\textbf{Top Ten Words}} & \textbf{Possible Topic} \\
\hline
Topic 1 & pt care resp vent respiratory secretions remains intubated abg plan psv bs support settings cont placed changes note wean rsbi coarse cpap continue peep suctioned clear extubated rr mask weaned & Respiratory failure/infection\\
\hline
Topic 2 & family pt ni care patient dnr stitle dr home daughter support team meeting wife son comfort note social doctor sw dni known time status hospital contact pt's work plan lastname & Discussion of end-of-life care\\
\hline
Topic 3 & hr resp gi pt cont gu neuro bs cv id note abd soft bp today stool social noted progress clear remains nursing skin urine sats foley npn yellow stable ls & Multiple physiological changes\\
\hline
Topic 4 & pain pt assessment response action plan control continue given dilaudid monitor chronic acute morphine iv po prn patient pca hr meds bp drain cont nausea ordered relief sbp pericardial assess & Assessments of patient responsiveness\\
\hline
Topic 10 & pt intubated vent propofol sedation sedated fentanyl peep tube versed secretions abg wean remains continue ett suctioned plan ps increased extubation settings ac sounds min cpap sputum respiratory hr ogt & Continued need for ventilation\\
\hline
Topic 38 & ml dl mg pm meq assessed icu ul total medications systems review pulse labs balance comments code hour rr min respiratory rhythm prophylaxis admission allergies blood urine mmhg status dose & Many labs tested\\
\hline
Topic 48 & ed pt patient transferred hospital pain admitted denies admission days nausea received ago presented micu showed vomiting past reports history given blood bp old year arrival known osh diarrhea unit & Emergency admission/transfer patient \\
\hline
\end{tabular}\label{tab:top_words}
\end{table}
%Note that our Topic 1 is similar to Topic 7 and 15 from \cite{ghassemi2014unfolding}, which were found to be associated with hospital mortality. 
\end{document}